\documentclass[10pt,twocolumn,letterpaper]{article}

\usepackage{iccv}
\usepackage{times}
\usepackage{epsfig}
\usepackage{graphicx}
\usepackage{amsmath}
\usepackage{amssymb}
\usepackage{multirow}
\usepackage{color}

\usepackage[breaklinks=true,bookmarks=false]{hyperref}

\iccvfinalcopy % *** Uncomment this line for the final submission

\def\iccvPaperID{****} % *** Enter the ICCV Paper ID here

\def\VSpaceValue{-1pt}%{-12pt}
\def\VSpaceValueEq{0pt}%{-8pt}
\def\VSpaceValueEqBottom{0pt}%{-5pt}

\begin{document}

%%%%%%%%% TITLE
\title{RGB-Infrared Cross-Modality Person Re-Identification  \\ via Joint Pixel and Feature Alignment}

\author{
Guan'an Wang$^{1,2}$ \ \ \ \ Tianzhu Zhang$^4$  \ \ \ \ Jian Cheng$^{1,2,3}$  \ \ \ \ Si Liu$^5$  \ \ \ \ Yang Yang$^{1}$   \ \ \ \  \footnotemark[2] Zengguang Hou$^{1,2,3}$  \\
$^1$Institute of Automation, Chinese Academy of Sciences, Beijing, China \\
$^2$University of Chinese Academy of Sciences, Beijing, China \\
$^3$Center for Excellence in Brain Science and Intelligence Technology, Beijing, China \\
$^4$University of Science and Technology of China, Beijing, China \  $^5$Beihang University, Beijing, China \\
\normalsize{
\{wangguanan2015, zengguang.hou\}@ia.ac.cn, tzzhang@ustc.edu.cn,
\{jcheng, yang.yang\}@nlpr.ia.ac.cn, liusi@buaa.edu.cn}
}

\maketitle

\renewcommand{\thefootnote}{\fnsymbol{footnote}}
\footnotetext[2]{Corresponding Author}

%%%%%%%%% ABSTRACT
\begin{abstract}
RGB-Infrared (IR) person re-identification is an important and  challenging task due
to large cross-modality variations between RGB and IR images.
Most conventional approaches  aim to bridge the cross-modality
gap with feature alignment by feature representation learning.
Different from existing methods, in this paper,
we propose a novel and end-to-end  Alignment 
Generative Adversarial Network (AlignGAN)
for the  RGB-IR RE-ID task.
The proposed   model   enjoys several merits. 
First, it can  exploit  pixel alignment and feature alignment
jointly. To the best of our knowledge,
this is the first work to model the two alignment strategies
jointly for the  RGB-IR RE-ID problem.
Second,  the proposed model 
consists of a pixel generator,
a feature generator and a joint discriminator. 
By playing a min-max game among the three components, 
our model is able to not only alleviate the cross-modality 
and intra-modality variations, 
but also learn identity-consistent features.
Extensive experimental results on two standard 
benchmarks demonstrate that the proposed  model 
performs favorably against state-of-the-art methods.
Especially, on SYSU-MM01 dataset, our model can 
achieve an  absolute gain of 
$15.4\%$ and $12.9\%$    in terms of   Rank-1 and mAP.
Code is released on \textit{https://github.com/wangguanan/AlignGAN}.
\end{abstract}

%%%%%%%%% BODY TEXT
\section{Introduction}
Person re-identification (Re-ID) is an important task in video surveillance, which aims to match pedestrian images of a person across disjoint camera views \cite{gong2014person}.
Its key challenges lie in large intra-class and small inter-class variations caused by different poses, illuminations, views, and occlusions.
To handle these issues, a large number of models for Re-ID problem have been proposed including hand-crafted descriptors \cite{ma2014covariance,yang2014salient,liao2015person}, metric  learning models \cite{zheng2013reidentification,koestinger2012large,liao2015efficient} and deep learning algorithms \cite{zheng2016person,hermans2017defense,sun2018beyond}.
Most of existing methods are focusing on visible cameras and formulate
the Re-ID task as a single-modality matching problem (RGB-RGB),
\textit{i.e.}, given a query image/video and match it against a set of gallery images/videos.

\begin{figure}[t]
\center
\includegraphics[scale=0.38]{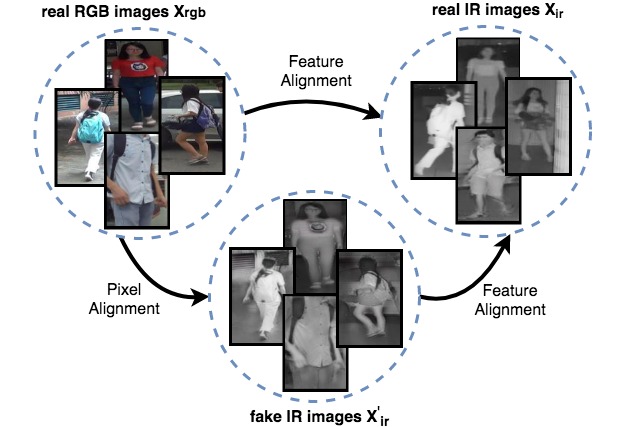}
\caption{
Most existing methods solve the RGB-IR Re-ID task via feature alignment 
by feature representation learning.
Different from existing methods,  our goal is to generate
fake IR images based on real RGB images via a pixel 
alignment module, and then match the generated fake IR images
and real IR images via a feature alignment module.
For more details, please refer to the text.
(Please view in color.)
}
\label{fig:simple_framework}
\vspace{\VSpaceValue}
\end{figure}

However, the visible cameras may not be able to capture valid appearance information
under poor illumination environments (e.g., at night),
which limits the applicability in
practical surveillance applications~\cite{wu2017rgb}.
In such case, imaging devices without relying
on visible light should be applied.
In many applications, the surveillance cameras could be heterogeneous, such
as near-infrared (IR), thermal and depth cameras.
Especially, most surveillance cameras can automatically switch from RGB to IR mode, which facilitates such cameras to work at night.
Thus, it is necessary to study the RGB-IR cross-modality
matching problem in real-world scenarios.
However, very few works have paid attention to the
Re-ID between RGB cameras and infrared cameras due to
the great differences between the two modalities.
As shown in Figure~\ref{fig:simple_framework},
RGB and IR images are intrinsically distinct and heterogeneous, and
have different wavelength ranges.
Here, RGB images  have three channels containing
colour information of visible light, while IR images have one channel containing information of invisible light.
As a result, even human can hardly recognize the persons well
by using colour information.

To deal with the above issues, existing cross-modality
re-id methods \cite{wu2017rgb,ye2018hierarchical,ye2018visible,dai2018cross,hao2019hsme} mainly focus on bridging the gap between
the RGB and IR images via feature alignment as shown in Figure~\ref{fig:simple_framework}.
The basic idea is to match real RGB and IR images via feature representation learning.
Due to the large cross-modality variation between two modalities, it is difficult to match  RGB and IR images directly in a shared feature space.
As shown in  Table~\ref{table:single_vs_cross},
we report the Rank-1, mAP scores and intra-class cosine similarity (ICCS) of the cmGAN~\cite{dai2018cross} (one state-of-the-art RGB-IR Re-ID model)  under single-modality and cross-modality  settings on the SYSU-MM01 dataset.
Note that larger ICCS value means higher similarity.
The results show the cmGAN performs much worse under the cross-modality setting and
cannot overcome the cross-modality variation well.

\begin{table}
\small
\center
\caption{
The results of cmGAN~\cite{dai2018cross} under different settings on SYSU-MM01 dataset in terms of Rank-1,  mAP,  and intra-class cosine similarity (ICCS).
}
\begin{tabular}{c|cc|cc}
\hline\hline
Settings & \multicolumn{2}{c|}{Single-Modality} &  \multicolumn{2}{|c}{Cross-Modality}\\
(query2gallery) & rgb2rgb & ir2ir & ir2rgb & rgb2ir \\
\hline
Rank-1  & 90.0 & 68.3 & 27.9 & 31.9 \\
mAP & 76.6 & 49.6 & 24.5 & 25.5 \\
ICCS & 0.892 & 0.879 & 0.701 & 0.701  \\
\hline
\hline
\end{tabular}
\label{table:single_vs_cross}
\vspace{\VSpaceValue}
\end{table}

\begin{figure}[!t]
\center
\includegraphics[scale=0.11]{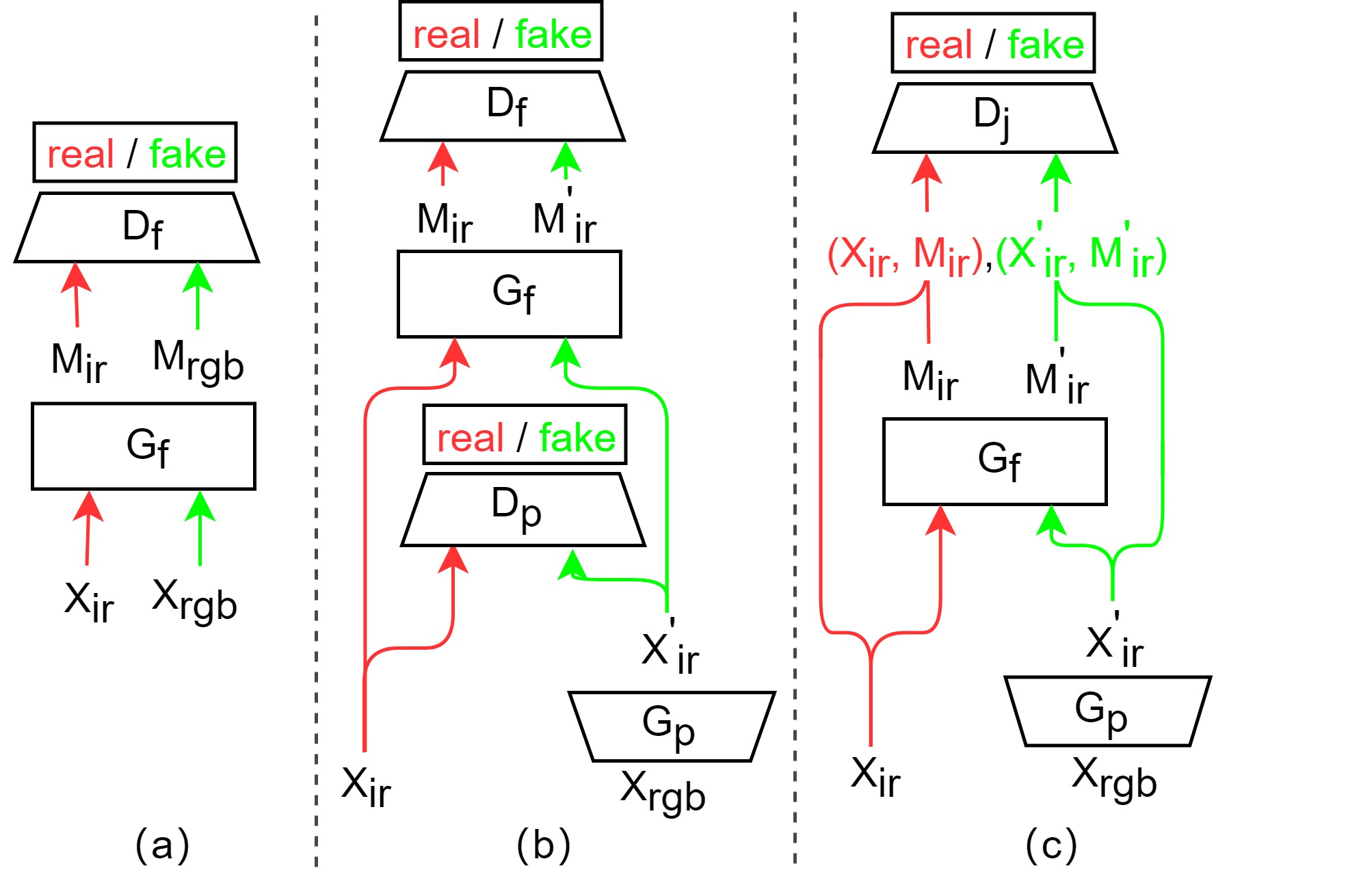}
\caption{Comparison of different alignment strategies. (a) Feature alignment. (b) Pixel and feature alignments with two discriminators. (c)  Pixel and feature alignments with a joint discriminator.}
\label{fig:compared_gans}
\vspace{\VSpaceValue}
\end{figure}

Different from the existing approaches by matching RGB and IR images straightly, a
heuristic method is to generate fake IR images based on real RGB images via a pixel alignment module, and then match the generated fake IR images and real IR images via a feature alignment module as shown in Figure~\ref{fig:simple_framework}.
The generated fake IR images are adopted to bridge the gap between
the RGB and IR images.
This basic idea can be achieved by using the model in Figure~\ref{fig:compared_gans}(b).
Here, the model consists of a pixel generator $G_p$ and a feature generator $G_f$ to align two modalities in the pixel and feature spaces, respectively. Correspondingly, the two generators are separately trained with two discriminators $D_p$ and $D_f$.
Thanks to the $G_p$  and $D_p$, fake IR images
can be generated to alleviate
the cross-modality variation in the pixel space.
Although the generated fake IR images look similar to real IR images, there are still
large intra-class discrepancies due to
viewpoint changes, pose variations and occlusions.
To overcome this issue, the  $G_f$  and $D_f$ are adopted.
Therefore, this model is designed for the RGB-IR cross-modality Re-ID by using pixel alignment and feature alignment, which is different from the model in
Figure~\ref{fig:compared_gans}(a) only using feature alignment.
However, the two alignment strategies are adopted separately, and
they may be not able to complement and enhance each other well
to obtain identity-consistent features.
This is because in Re-ID task labels of training and test set are unshared. Aligned features cannot maintain identity-consistency by fitting labels in training set. For example, person A may be aligned to person B.

Inspired by the above discussions, in this paper,
we propose a novel Alignment Generative Adversarial Network (AlignGAN) to simultaneously alleviate the cross-modality variation in the pixel space, the intra-modality variation in the feature space, and maintain the identity-consistent features for the RGB-IR cross-modality Re-ID task.
As shown in Figure~\ref{fig:compared_gans}(c),
to reduce the cross-modality variation, we adopt a pixel generator $G_p$ to generate identity-maintained fake IR images based on RGB images.
Then, to alleviate the intra-modality variation, we use a feature generator $G_f$ to encode fake and real IR images to a shared feature space by utilizing the identity-based classification and triplet losses.
Finally, to learn identity-consistent features, we propose a joint discriminator $D_j$ by making $G_p$ and $G_f$ learn from each other.
Here, $D_j$ takes image-feature pairs as inputs and classifies real IR image-feature pairs with the same identity as real, and the others as fake. Correspondingly,   $G_{p}$ and $G_{f}$ are optimized to fool  $D_j$. Thus, negative pairs from different identities can be penalized and the aligned features are explicitly forced to maintain identity with corresponding images.
By playing a min-max game between $D_j$ and $G_p+G_f$,  both cross-modality and intra-modality variation can be reduced, meanwhile the identity-consistent features can be learned.

The major contributions of this work can be summarized as follows.
(1) We propose a novel Alignment Generative Adversarial Network for the  RGB-IR RE-ID task
by exploiting pixel alignment and feature alignment  jointly. To the best of our knowledge,
this is the first work to model the two alignment strategies
jointly for the  RGB-IR RE-ID problem.
(2) The proposed model consists of a pixel generator, a feature generator and a joint discriminator. By playing a min-max game among the three components, our model is able to not only reduce the cross-modality and intra-modality variations, but also learn identity-consistent features.
(3) Extensive experimental results on two standard
benchmarks demonstrate that the proposed  model
performs favorably against state-of-the-art methods.
Especially, on SYSU-MM01 dataset, our model
achieves a significant improvement of
$15.4\%$ Rank-1 and $12.9\%$ mAP, respectively.

\section{Related Works}

%In this section, we briefly overview methods that are related to RGB-RGB person re-identification,  RGB-IR person re-identification, and generative adversarial networks.

\textbf{RGB-RGB Person Re-Identification.} RGB-RGB person re-identification addresses the problem of matching pedestrian RGB images across disjoint visible cameras \cite{gong2014person}, which is widely used in video surveillance, public security and smart city, can also be used to improve tracking \cite{zhang2019robust,zhang2018correlation,zhang2019learning}. 
The key challenges lie in the large intra-class variation caused by different views,  poses, illuminations, and occlusions.
Existing methods can be grouped into hand-crafted descriptors \cite{ma2014covariance,yang2014salient,liao2015person, yang2016large}, metric learning methods \cite{zheng2013reidentification,koestinger2012large,liao2015efficient} and deep learning algorithms  \cite{zheng2016person,hermans2017defense,sun2018beyond,luo2019bag,zheng2019joint,guan2019color,wu2019clustering,yang2017unsupervised,yang2019cn,li2018harmonious,li2017person,chen2017person,wang2018transferable}.
The goal of hand-crafted descriptors is to design robust features. For example,
Yang \textit{et al.} \cite{yang2014salient} explore color information by using salient color names.
Metric learning methods are designed to make a pair of true matches have a relatively smaller distance than that of a wrong match pair in a  discriminant manner.
Zheng \textit{et al.} \cite{zheng2013reidentification} formulate person RE-ID
as a relative distance comparison  learning problem in order to
learn the optimal similarity measure between a pair of person images.
Deep learning algorithms adopt deep neural networks to straightly learn robust and discriminative features in an end-to-end manner. 
For example, \cite{zheng2016person,hermans2017defense} learn identity-discriminative features by fine-tuning a pre-trained CNN to minimize a classification loss or a triplet loss.
Most of exiting methods focus on the RGB-RGB Re-ID task, and cannot perform well for the RGB-IR Re-ID task, which limits the applicability in practical surveillance scenarios.

\textbf{RGB-IR Person Re-Identification.}
RGB-IR Person re-identification  attempts to match RGB and IR images of a person under disjoint cameras. Besides the difficulties of RGB-RGB Re-ID, RGB-IR Re-ID faces a new challenge due to cross-modality variation between RGB and IR images.
In~\cite{wu2017rgb}, Wu \textit{et al.}  collect a cross-modality RGB-IR dataset named SYSU RGB-IR Re-ID. The proposed method explores three different network structures and uses deep zero-padding for training one-stream network towards automatically evolving domain-specific nodes in the network for cross-modality matching.
Ye \textit{et al.}~\cite{ye2018hierarchical,ye2018visible} propose modality-specific and modality-shared metric losses and a new bi-directional dual-constrained top-ranking loss to learn discriminative feature representations for RGB-IR Re-ID.
In~\cite{dai2018cross},  Dai \textit{et al.} introduce a  cross-modality generative adversarial network (cmGAN)  to reduce the distribution divergence of RGB and IR features.
Very recently, Hao \textit{et al.}~\cite{hao2019hsme} achieve visible thermal person re-
identification via a hyper-sphere manifold embedding model.
Most of the above methods  mainly focus on bridging the gap between
RGB and IR images via feature alignment, which ignores
the large cross-modality variation in the pixel space.
Different from these methods, our proposed model performs pixel alignment and feature alignment jointly, which is able to
not only reduce the cross-modality and intra-modality variations, but also learn identity-consistent features.

\textbf{Generative Adversarial Networks.}
Generative Adversarial Network (GAN) \cite{goodfellow2014generative} learns data distribution in a self-supervised way via the adversarial training, which has been widely used in  image translation \cite{isola2017image,zhu2017unpaired,choi2018stargan} and  domain adaptation \cite{ganin2016domain,hoffman2018cycada,dou2018unsupervised,shao2018feature}.
Pix2Pix \cite{isola2017image}, CycleGAN \cite{zhu2017unpaired} and StarGAN \cite{choi2018stargan} learn image translations between two or multi domains.
However, those works only focus on image translation, which cannot be used for RGB-IR Re-ID, a cross-modality matching task. Recently, some GAN based domain adaptation methods are proposed.
DANN \cite{ganin2016domain} and Seg-CT-UDA \cite{dou2018unsupervised} minimizes adversarial discriminator accuracy to reduce the distribution divergence between source and target features. However, they only focus on aligning features between two domains, while fails to deal with variations of images.
CyCADA \cite{hoffman2018cycada} uses two GAN models to generate images and features for segmentation, HADDA \cite{shao2018feature} use reconstructed images to constrain features. However, both assume that training and test data should have the same class labels, which is not hold for person re-id.
Furthermore, our pixel and feature alignment modules are jointly learned in an unified GAN framework, and  identity-consistent features can be obtained by making the two modules learn from each other.

\section{Alignment Generative Adversarial Network}

\begin{figure*}
\center
\includegraphics[scale=0.080]{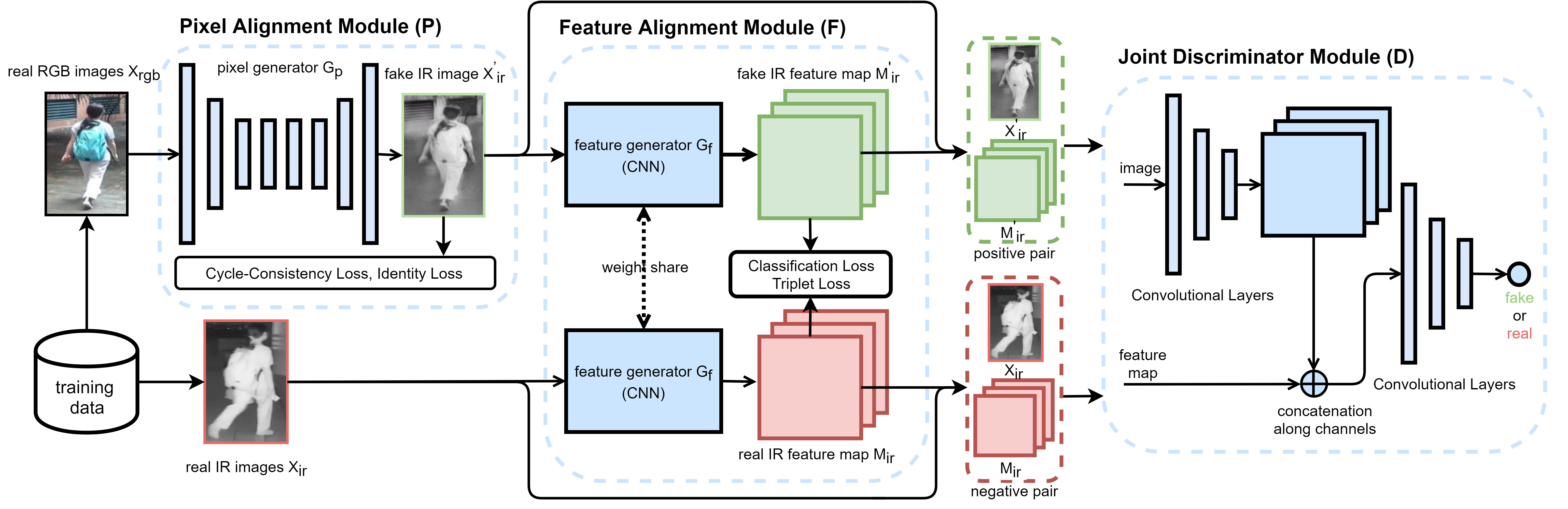}
\caption{
Framework of the proposed AlignGAN model. It consists of a pixel alignment module ($\mathcal{P}$), a feature alignment module ($\mathcal{F}$), and a joint discriminator module ($\mathcal{D}$). 
The $\mathcal{P}$ can generate fake IR images $X^{'}_{ir}$ to alleviate  the cross-modality variation, the $\mathcal{F}$ can alleviate the intra-modality variation, and the $\mathcal{D}$ can obtain identity-consistent features by making $\mathcal{P}$ and $\mathcal{F}$ learn from each other and penalizing negative pairs which are not real or belong to different identities.
}
\label{fig:framework}
\vspace{\VSpaceValue}
\end{figure*}

In this section, we introduce the details of the proposed Alignment Generative Adversarial Network (AlignGAN) for  the RGB-IR Re-ID.
As shown in Figure \ref{fig:framework}, our AlignGAN consists of a pixel alignment module ($\mathcal{P}$), a feature alignment module ($\mathcal{F}$) and a joint discriminator module ($\mathcal{D}$).
The $\mathcal{P}$ reduces the cross-modality variation by translating RGB images to identity-maintained fake IR images.
The $\mathcal{F}$ alleviates the intra-modality variation by encoding fake and real IR images to a shared feature space by minimizing the identity-based classification and triplet losses.
The $\mathcal{D}$ enforce the identity-consistent features by making the $\mathcal{P}$ and $\mathcal{F}$ learn from each other and penalizing negative pairs which are not real or belong to different identities.
By playing a min-max game between the $\mathcal{D}$ and the $\mathcal{D}+\mathcal{F}$, both cross-modality and intra-modality variations can be reduced, meanwhile identity-consistent features can be learned.

\subsection{Pixel Alignment Module\label{section:image}}

% pixel generator
As shown in Figure \ref{fig:simple_framework}, there is a large cross-modality variation between RGB and IR images, which significantly increases the difficulties of the RGB-IR Re-ID task.
To reduce the cross-modality variation, we propose to translate real RGB images $X_{rgb}$ to fake IR images $X^{'}_{ir}$, which have the IR style and maintain their original identities. The generated fake IR images $X^{'}_{ir}$ can be used to bridge the gap between RGB and IR images.
To this end, we introduce a pixel generator $G_{p}$ which learns a mapping from RGB images $X_{rgb}$ to IR images $X_{ir}$, and train it to produce fake IR images $X^{'}_{ir}=G_{p}(X_{rgb})$ that fool a discriminator.
Conversely, the discriminator attempts to distinguish real IR images $X_{ir}$ from fake ones $X^{'}_{ir}$. By playing the min-max game as in  \cite{goodfellow2014generative}, the proposed model can make the fake IR images $X^{'}_{ir}$ as realistic as possible. We mark the loss as $\mathcal{L}_{gan}^{pix}$ and define it in Eq.(\ref{loss:gan_image}).

% cycle-consistency loss
Although $\mathcal{L}_{gan}^{pix}$ ensures fake IR images $X^{'}_{ir}$ will resemble data drawn from real IR images $X_{ir}$, there is no way to guarantee that $X^{'}_{ir}$ preserves the structure or content of their original RGB images $X_{rgb}$.
To handle this issue, as in \cite{zhu2017unpaired}, we introduce a cycle-consistency loss.
Specifically, a mapping from IR images to RGB ones $G_{p^{'}}$ is trained with a GAN model as in \cite{zhu2017unpaired}. For simplification, we don't show its definition in the loss.
Then, we require that mapping a RGB(IR) image to a IR(RGB) one and back to a RGB(IR) one reproduces the original one.
The cycle-consistency loss $\mathcal{L}_{cyc}$ can be formulated as below:
\vspace{\VSpaceValueEq}
\begin{equation}
\begin{aligned}
\mathcal{L}_{cyc} & =  ||G_{p^{'}}(G_{p}(X_{rgb})) - X_{rgb}||_{1} \\
& + ||G_{p}(G_{p^{'}}(X_{ir})) - X_{ir}||_{1}
\end{aligned}
\end{equation}
\vspace{\VSpaceValueEq}

% maintain identites of fake images
Additionally, we make fake IR images $X^{'}_{ir}$ maintain identities of corresponding RGB images $X_{rgb}$ from two aspects.
On one hand, $X^{'}_{ir}$ should be classified to the identities of the corresponding $X_{rgb}$. On the other hand, $X^{'}_{ir}$ should satisfy a triplet constraint supervised by the identities of corresponding $X_{rgb}$.
We mark the two losses as $\mathcal{L}^{pix}_{cls}$ and $\mathcal{L}^{pix}_{tri}$ and formulate them in Eq.(\ref{loss:image_identi_consistency}), respectively.
\vspace{\VSpaceValueEq}
\begin{equation}
\begin{aligned}
\mathcal{L}^{pix}_{cls} & =  \mathcal{L}_{cls}(X^{'}_{ir})  =  E_{x \in X^{'}_{ir}}[-log \ p(x)]\\
\mathcal{L}^{pix}_{tri} & =  \frac{1}{2} [\mathcal{L}_{tri}(X^{'}_{ir}, X_{ir}, X_{ir}) + \mathcal{L}_{tri}(X_{ir}, X^{'}_{ir}, X^{'}_{ir})]
\end{aligned}
\label{loss:image_identi_consistency}
\end{equation}
Here, $p(\cdot)$ is the predicted probability of the input belonging the ground-truth identity. The ground-truth identities of fake IR images $X^{'}_{ir}$ are the same with ones of corresponding original RGB images $X_{rgb}$.
$\mathcal{L}_{tri}$ is defined in Eq.(\ref{loss:triplet_loss}), where $x_a$ and $x_p$ are a positive pair belonging to the same identity, $x_a$ and $x_n$ are a negative pair belonging to different identities, $D_{x_1, x_2}$ is the cosine distances between $x_1$ and $x_2$ in the embedding space of the embedder, $m$ is a margin parameter and empirically set $1.0$, and $[x]+=max(0,x)$.
\vspace{\VSpaceValueEq}
\begin{equation}
\begin{aligned}
& \mathcal{L}_{tri}(X_1, X_2, X_3) =E\{ [m - D_{x_a, x_p} + D_{x_a, x_n}]_+\} \\
& s.t. \ x_a \in X_1, x_p \in X_2, x_n \in X_3.
\end{aligned}
\label{loss:triplet_loss}
\end{equation}
\vspace{\VSpaceValueEq}

In summary, the overall loss of our pixel alignment module is shown in Eq.(\ref{loss:image_alignment_module}), where $\lambda_{cyc}$ and $\lambda^{pix}_{id}$ are weights of the corresponding terms, $\lambda_{cyc}$ is set 10 as in \cite{zhu2017unpaired} and $\lambda^{pix}_{id}$ is set 1.0 via cross-validation.
\vspace{\VSpaceValueEq}
\begin{equation}
\mathop{min}\limits_{\mathcal{P}}\mathcal{L}^{pix} = \mathcal{L}^{pix}_{gan} + \lambda_{cyc}\mathcal{L}_{cyc} + \lambda^{pix}_{id}(\mathcal{L}^{pix}_{cls}+\mathcal{L}^{pix}_{tri})
\label{loss:image_alignment_module}
\end{equation}
\vspace{\VSpaceValueEq}

\subsection{Feature Alignment Module\label{section:feature}}

Although the pixel alignment module reduces the cross-modality variation, there is still a large intra-modality variation caused by different poses, views, illuminations and so on. To overcome it, we propose a feature alignment module $\mathcal{F}$, where a feature generator $G_f$ encodes fake IR images $X^{'}_{ir}$ and real IR images $X_{ir}$ to a shared space by minimizing identity-based classification \cite{zheng2016person} and triplet losses \cite{hermans2017defense}.
Specifically, we adopt a CNN as the feature generator $G_f$ to learn feature maps $M$ and then averagely pool them to feature vectors $V$. $G_f$ takes $X_{ir}$ and $X^{'}_{ir}$ as inputs and is optimized with the classification loss $\mathcal{L}^{feat}_{cls}$ of a classifier and the triplet loss $\mathcal{L}^{feat}_{tri}$ of an embedder as below:
\vspace{\VSpaceValueEq}
\begin{equation}
\begin{aligned}
\mathcal{L}^{feat}_{cls}&  = \mathcal{L}_{cls}(X_{ir}\cup X^{'}_{ir}) = E_{x \in X_{ir} \cup X^{'}_{ir}}[-log \ p(x)] \\
\mathcal{L}^{feat}_{tri} & = \mathcal{L}_{tri}(X_{ir}, X^{'}_{ir}, X^{'}_{ir}) + \mathcal{L}_{tri}(X^{'}_{ir}, X_{ir} X_{ir})
\end{aligned}
\label{loss:feature_identi_loss}
\vspace{\VSpaceValueEqBottom}
\end{equation}
where $\cup$ means set union,  $p(\cdot)$ is the predicted probability of the input belonging the ground-truth identity. The ground-truth identities of $X^{'}_{ir}$ are the same with ones of corresponding $X_{rgb}$,
$\mathcal{L}_{tri}$ is defined in Eq.(\ref{loss:triplet_loss}).

Although the classification and triplet losses in Eq.(\ref{loss:feature_identi_loss}) can learn identity-aware features, they cannot deal with the modality-variation in the feature space well, which limits the accuracy of the RGB-IR RE-ID. To solve it, we further adopt a GAN loss in the feature space to remit the cross-modality variation by reducing the distribution divergence. Specifically, we use a discriminator to distinguish feature maps of real IR images $M_{ir}$ from ones of fake IR images $M^{'}_{ir}$. Contrarily, the feature generator $G_{f}$ is optimized to fool the discriminator. By playing the min-max game as in \cite{goodfellow2014generative}, the distribution divergence between $M^{'}_{ir}$ and $M_{ir}$ can be reduced. The detailed formulation of the GAN loss $\mathcal{L}_{gan}^{feat}$ is described in Eq.(\ref{loss:gan_feature}).

Thus, the overall loss of our feature alignment module can be formulated as in Eq.(\ref{loss:feature_alignment_module}), where $\lambda_{gan}^{feat}$ is the weight of the GAN loss and set via 0.1 cross-validation.
\vspace{\VSpaceValueEq}
\begin{equation}
\mathop{min}\limits_{\mathcal{F}}\mathcal{L}^{feat} = \mathcal{L}^{feat}_{cls} + \mathcal{L}^{feat}_{tri} + \lambda_{gan}^{feat} \mathcal{L}_{gan}^{feat}
\label{loss:feature_alignment_module}
\end{equation}
\vspace{\VSpaceValueEq}

\subsection{Joint Discriminator Module \label{section:discriminator}}

Our joint discriminator module consists of a joint discriminator ($D_j$), which takes an image-feature pair $(X,M)$ as a input and outputs one logit, where 1 means real and 0 fake.  Only the pairs of real IR images $X_{ir}$ and real IR features $M_{ir}$ with the same identity are classified to real, and the others fake. Thus, when optimizing the pixel alignment module ($\mathcal{P}$) and the feature alignment module ($\mathcal{F}$) to fool the joint discriminator module ($\mathcal{D}$), there are two advantages. Firstly, by playing the min-max game,  fake IR images $X^{'}_{ir}$ will be realistic and fake IR features $M^{'}_{ir}$ will have similar distribution of real IR features $M_{ir}$.

Secondly, $M^{'}_{ir}$ can maintain the identity-consistency with the corresponding image $X^{'}_{ir}$.
This is because our AlignGAN plays like a conditional GAN (cGAN) \cite{mirza2014conditional}. In the cGAN, the classes of generated data will depend  on the input conditions. Here, the input images of $\mathcal{F}$ act as the condition, and the classes of learned features will be related to those images.
To this end, the objective function of the $\mathcal{D}$ can be formulated as in Eq.(\ref{loss:joint_discriminator_module}):
\vspace{\VSpaceValueEq}
\begin{equation}
\mathop{min}\limits_{\mathcal{D}}\mathcal{L}^{D} = \mathcal{L}^{D}_{real} + \mathcal{L}^{D}_{fake}
\label{loss:joint_discriminator_module}
\end{equation}
\vspace{\VSpaceValueEq}
\begin{equation}
\begin{aligned}
\mathcal{L}^{D}_{real} & = E_{(x,m)\in(X_{ir},M_{ir})}[log \ D_j(x,m)] \\
\mathcal{L}^{D}_{fake} & = E_{(x,m) \in (\widetilde{X}_{ir}, \widetilde{M}_{ir}) + (\overline{X}_{ir}, \overline{M}_{ir})}[log \ 1-D_j(x,m)]
\end{aligned}
\end{equation}
where $x$ and $m$ of $(X_{ir}, M_{ir})$ are both real and belong to the same identity, $x$ and $m$ of $(\widetilde{X}_{ir}, \widetilde{M}_{ir})$ belong to the same identity and at least one of them is fake, $x$ and $m$ of $(\overline{X}_{ir}, \overline{X}_{ir})$ are both real but belong  to different identities.

Correspondingly, to fool the joint discriminator module , the GAN losses of the pixel and feature alignment modules can be formulated in Eq.(\ref{loss:gan_image}) and Eq.(\ref{loss:gan_feature}), respectively.
\vspace{\VSpaceValueEq}
\begin{equation}
\mathcal{L}_{gan}^{pix} = E_{(x,m)\in(X^{'}_{ir}, M_{ir}) \cup (X^{'}_{ir}, M^{'}_{ir})}[log D_j(x,m)]
\label{loss:gan_image}
\end{equation}
\vspace{\VSpaceValueEqBottom}
\vspace{\VSpaceValueEq}
\begin{equation}
\mathcal{L}_{gan}^{feat} = E_{(x,m)\in(X_{ir},M^{'}_{ir}) \cup (X^{'}_{ir}, M^{'}_{ir})}[log D_j(x,m)]
\label{loss:gan_feature}
\end{equation}
where $x$ and $m$ of $(X^{'}_{ir}, M_{ir})$ belong to the same identity, $x$ is fake and $m$ is real. $x$ and $m$ of $(X_{ir},M^{'}_{ir})$ belong to the same identity, $x$ is real and $m$ is fake. Similarly, $x$ and $m$ of $(X^{'}_{ir},M^{'}_{ir})$ belong to the same identity and are both fake.

\begin{table*}[!t]
\small
\center
\caption{Comparison with the state-of-the-arts on SYSU-MM01 dataset. The R1, R10, R20 denote Rank-1, Rank-10 and Rank-20 accuracies (\%), respectively. The mAP denotes mean average precision score (\%).
}
\setlength{\tabcolsep}{5.0pt}
\begin{tabular}{c|cccc|cccc|cccc|cccc}
\hline\hline
\multicolumn{1}{c}{\multirow{3}{*}{Methods}} & \multicolumn{8}{|c}{\textit{All-Search}} & \multicolumn{8}{|c}{\textit{Indoor-Search}} \\
\cline{2-17}
 &  \multicolumn{4}{|c}{\textit{Single-Shot}} &  \multicolumn{4}{|c|}{\textit{Multi-Shot}} &  \multicolumn{4}{c|}{\textit{Single-Shot}} &  \multicolumn{4}{c}{\textit{Multi-Shot}} \\
~ & R1 & R10 & R20 & mAP  & R1 & R10 & R20 & mAP  & R1 & R10 & R20 & mAP  & R1 & R10 & R20 & mAP \\
\hline
HOG & 2.76 & 18.3 & 32.0 & 4.24 &   3.82 & 22.8 & 37.7 & 2.16 &  3.22 & 24.7 & 44.6 & 7.25 &   4.75 & 29.1 & 49.4 & 3.51 \\
LOMO & 3.64 & 23.2 & 37.3 & 4.53   & 4.70 & 28.3 & 43.1 & 2.28   & 5.75 & 34.4 & 54.9 & 10.2 &   7.36 & 40.4 & 60.4 &  5.64 \\
Two-Stream &   11.7 & 48.0 & 65.5 & 12.9 &   16.4 & 58.4 & 74.5 & 8.03 &    15.6 & 61.2 & 81.1 & 21.5 &    22.5 & 72.3 & 88.7 & 14.0 \\
One-Stream &  12.1 &  49.7 & 66.8 & 13.7 &   16.3 & 58.2 & 75.1 & 8.59 &   17.0 & 63.6 & 82.1 & 23.0 &    22.7 & 71.8 & 87.9 & 15.1 \\
Zero-Padding &  14.8 & 52.2 & 71.4 & 16.0 &   19.2 & 61.4 & 78.5 & 10.9 &  20.6  & 68.4 & 85.8 & 27.0 &  24.5 & 75.9 & 91.4 & 18.7 \\
BCTR & 16.2 & 54.9 & 71.5 & 19.2 &     - & - & - & - &     - & - & - & - &     - & - & - & - \\
BDTR & 17.1 & 55.5 & 72.0 & 19.7 &     - & - & - & - &     - & - & - & - &     - & - & - & - \\
D-HSME & 20.7 & 62.8 & 78.0 & 23.2 &     - & - & - & - &     - & - & - & - &     -  & - & - & - \\
cmGAN & 27.0 & 67.5 & 80.6 & 27.8   & 31.5 & 72.7 & 85.0 & 22.3   & 31.7 & 77.2 & 89.2 & 42.2 &  37.0 & 80.9 & 92.3 & 32.8 \\
\hline
{\textit{Ours}} & \textbf{42.4} & \textbf{85.0} & \textbf{93.7} & \textbf{40.7}   & \textbf{51.5} & \textbf{89.4} & \textbf{95.7} & \textbf{33.9}   & \textbf{45.9} & \textbf{87.6} & \textbf{94.4} & \textbf{54.3} &  \textbf{57.1} & \textbf{92.7} & \textbf{97.4} & \textbf{45.3}  \\
\hline
\hline
\end{tabular}
\label{table:state-of-the-art}
\vspace{\VSpaceValue}
\end{table*}

\subsection{Train and Test}

During the training stage, our AlignGAN can be trained by alternatively optimizing corresponding loss of each module in Eq.(\ref{loss:image_alignment_module}), Eq.(\ref{loss:feature_alignment_module}) and Eq.(\ref{loss:joint_discriminator_module}), respectively.
During the test stage, only the pixel alignment module $\mathcal{P}$ and the feature alignment module $\mathcal{F}$ are used. For IR images $X_{ir}$, we straightly use the $\mathcal{F}$ to learn feature vectors $V_{ir}$. For RGB images $X_{rgb}$, we first use the $\mathcal{P}$ translate to them to fake IR images $X^{'}_{ir}$, and then extract their feature vectors $V^{'}_{ir}$ using the $\mathcal{F}$. Finally, matching is conducted by computing cosine similarities of feature vectors between the probe images and gallery ones.

\section{Experiments}

%In this section, we perform extensive experiments to evaluate our model. Section \ref{section:dataset} and Section \ref{section:details} describe the experiment settings. Section \ref{section:results1} compares our model with the state-of-the-arts.  Section \ref{section:analysis} further analyzes our model including ablation study, parameters analysis and effect of CNN backbones. Section \ref{section:visualization} visualizes the outputs of pixel and feature alignment modules.

\subsection{Dataset and Evaluation Protocol \label{section:dataset}}

\textbf{Dataset}. 
We evaluate our model on two standard
benchmarks including  SYSU-MM01 and RegDB.
(1) SYSU-MM01 \cite{wu2017rgb} is a popular RGB-IR Re-ID dataset, which includes 491 identities from 4 RGB cameras and 2 IR ones.  The training set contains 19,659 RGB images and 12,792 IR images of 395 persons and the test set contains 96 persons.
Following \cite{wu2017rgb}, there are two test modes, \textit{i.e.} \textit{all-search} mode and \textit{indoor-search} mode.
For the \textit{all-search} mode, all images are used. For the \textit{indoor-search} mode, only indoor images from $1st, 2nd, 3rd, 6th$ cameras are used. For both modes, the \textit{single-shot} and \textit{multi-shot} settings are adopted, where 1 or 10 images of a person are randomly selected to form the gallery set. Both modes use IR images as probe set and RGB images as gallery set.
(2) RegDB \cite{nguyen2017person} contains 412 persons, where each person has 10 images from a visible camera and 10 images from a thermal camera.

\textbf{Evaluation Protocols}. The Cumulative Matching Characteristic (CMC) and mean average precision (mAP) are used as evaluation metrics.
Following \cite{wu2017rgb}, the results of SYSU-MM01 are evaluated with offical code based on the average of 10 times repeated random split of gallery and probe set. Following \cite{ye2018hierarchical,ye2018visible}, the results of RegDB are based the average of 10 times repeated random split of training and testing sets. Detailed settings can be found in github of corresponding author.

\subsection{Implementation Details \label{section:details}}

Following \cite{dai2018cross}, we adopt the ResNet-50 \cite{he2016deep} pre-trained on ImageNet \cite{russakovsky2015imagenet} as our CNN backbone, use its Pool5 layer as our feature map $M$ and averagely pool $M$ to obtain feature vector $V$.  For the classification loss, the classifier takes the feature vector $V$ as the input, includes a 256-dim fully-connected (FC) layer followed by batch normalization \cite{ioffe2015batch}, dropout \cite{russakovsky2015imagenet} and ReLU \cite{russakovsky2015imagenet} as the middle layer, and an FC layer with identity number logits as the output layer. The dropout rate is set 0.5 empirically. For the triplet loss, the embedder is an FC layer which maps a feature vector $V$ to a 256-dim embedding vector.

We implement our model with Pytorch. The training images are augmented with the horizontal flip. The batch size is set to 144 (18 persons, 4 RGB images and 4 IR images for a person). For learning rates, we set the classifier and the embedder as 0.2 and the imagenet pre-trained CNN part as 0.02 and optimize them via SGD. We set the learning rates of the pixel alignment and joint discriminator modules as 0.0002 and optimize them via the Adam \cite{radford2016unsupervised}. The learning rates are decayed by 0.1 after 5,000 iterations, and the model is trained for 10,000 iterations in total.

\subsection{Results on SYSU-MM01 Dataset \label{section:results1}}

We compare our model with 9 methods including hand-crafted features (HOG \cite{dalal2005histograms}, LOMO \cite{liao2015person}), feature learning with the classification loss (One-Stream, Two-Stream, Zero-Padding) \cite{wu2017rgb}, feature learning with both classification and ranking losses (BCTR, BDTR) \cite{ye2018hierarchical}, metric learning (D-HSME \cite{hao2019hsme}), and reducing distribution divergence of features (cmGAN \cite{dai2018cross}). The  experimental results are shown in Table \ref{table:state-of-the-art}.

From the view of the evaluation protocol, \textit{i.e.} \textit{all-search/indoor-search} and \textit{single-shot/multi-shot}, two phenomenons can be observed. Firstly, for the same method, \textit{indoor-search} performs better than \textit{all-search}. This is because images of indoor have less background-variation, which makes matching easier.
Secondly, we find that Rank scores of \textit{multi-shot} are higher than ones of \textit{single-shot}, but mAP scores of \textit{multi-shot} are lower than ones of \textit{single-shot}. This is because there are 10 images of a person in gallery set under the \textit{multi-shot} mode, but only one under the \textit{single-shot} mode. As a consequence, under the \textit{multi-shot} mode, it's much easier to hit an image but difficult to hit all images. This situation is inverse under the \textit{single-shot} mode. If not specified, we analyze each method under the \textit{single-shot\&all-search} mode below.

From the view of methodology,  several phenomenons can be observed.
Firstly, LOMO only achieves 3.64\%  and 4.53\% in terms of Rank-1 and mAP scores, respectively, which shows that hand-crafted features cannot be generalized to the RGB-IR Re-ID task.
Secondly, One-Stream, Two-Stream and Zero-Padding significantly outperform hand-crafted features by at least 8\%  and 8.3\% in terms of Rank-1 and mAP scores, respectively. This verifies that the classification loss contributes to learning identity-discriminative features.
Thirdly, BCTR and BDTR further improve Zero-Padding by 1.4\% in terms of Rank-1 and by 3.2\% in terms of mAP scores. This shows that the ranking and classification losses are complementary.
Additionally, D-HSME outperforms BDTR by 3.6\% Rank-1 and 3.5\% mAP scores, which demonstrates the effectiveness of metric learning.
In addition, cmGAN outperforms D-HSME by 6.3\% Rank1 and 4.6\% mAP scores, implying the effectiveness of adversarial training.
Finally, Our proposed AlignGAN significantly outperforms the state-of-the-art method by 15.4\%  and 12.9\% in terms of Rank-1 and mAP scores, which demonstrates the effectiveness of our model for the RGB-IR Re-ID task.

\begin{table}
\small
\center
\caption{
Comparison with state-of-the-arts on RegDB dataset under different query settings. 
Refer to the text for more details.
%ir2rgb means that IR images are query set and RGB images are gallery set, and so on.
}
\center
\begin{tabular}{c|cc|cc}
\hline\hline
\multirow{2}{*}{Methods} & \multicolumn{2}{|c|}{thermal2visible} &  \multicolumn{2}{|c}{visible2thermal} \\
~ & Rank-1 & mAP & Rank-1 & mAP \\
\hline
Zero-Padding & 16.7 & 17.9 & 17.8 & 31.9 \\
TONE & 21.7 & 22.3 & 24.4 & 20.1 \\
BCTR &  - & - & 32.7 & 31.0 \\
BDTR & 32.8 & 31.2 & 33.5 & 31.9 \\
D-HSME & 50.2 & 46.2 & 50.9 & 47.0 \\
\hline
$Basel.$ & 32.7 & 34.9 & 33.1 & 35.5\\ 
\textit{Ours} & 56.3 & 53.4 & 57.9 & 53.6 \\
\hline
\hline
\end{tabular}
\label{table:sota_regdb}
\vspace{\VSpaceValue}
\end{table}

\begin{table}[t]
\small
\center
\caption{Comparison with different variants of our AlignGAN on SYSU-MM01 dataset under the \textit{single-shot\&all-search} mode.}
\begin{center}
\setlength{\tabcolsep}{5.0pt}
\begin{tabular}{c|cccc}
\hline\hline
Methods & Rank-1 & Rank-10 & Rank-20 & mAP  \\
\hline
$Basel.$ & 29.6 & 74.9 & 86.1 & 33.0 \\
$PixAlign$ & 40.6 & 81.6 & 90.6 & 38.7 \\
$FeatAlign$ & 34.1 & 79.6 & 89.1 & 36.2 \\
$AlignGAN^-$ & 36.2 & 80.1 & 90.2 & 34.2 \\
$AlignGAN$ & \textbf{42.4} & \textbf{85.0} & \textbf{93.7} & \textbf{40.7}  \\
\hline
\hline
\end{tabular}
\end{center}
\label{table:ablation_study}
\vspace{\VSpaceValue}
\end{table}

\subsection{Results on RegDB Dataset \label{section:results2}}

We evaluate our model on RegDB dataset and compare it with Zero-Padding \cite{wu2017rgb}, TONE \cite{ye2018visible}, BCTR \cite{ye2018hierarchical}, BDTR \cite{ye2018visible} and $Basel.$. $Basel.$ is defined in Section \ref{section:analysis}, which learn thermal and visible images with classification and triplet losses.
We adopt visible2thermal and thermal2visible modes. Here, the visible2thermal means that visible images are query set and thermal images are gallery set, and so on.
As shown in Table \ref{table:sota_regdb}, our model can significantly outperform the state-of-the-arts by 23.5\% and 24.4\% in terms of Rank-1 scores with thermal2visible and visible2thermal modes, respectively.
Compared with HSME, our model outperforms it by $6.1\%/7.0\%$ Rank-1 scores in terms of  thermal2visible/visible2thermal modes on RegDB, and  
obtains an absolute gain of $21.72\%/17.58\%$ Rank-1 score on SYSU-MM01 Dataset. 
Overall, the results verify the effectiveness of our model.

\subsection{Model Analysis \label{section:analysis}}

\begin{figure}[t]
\center
\includegraphics[scale=0.45]{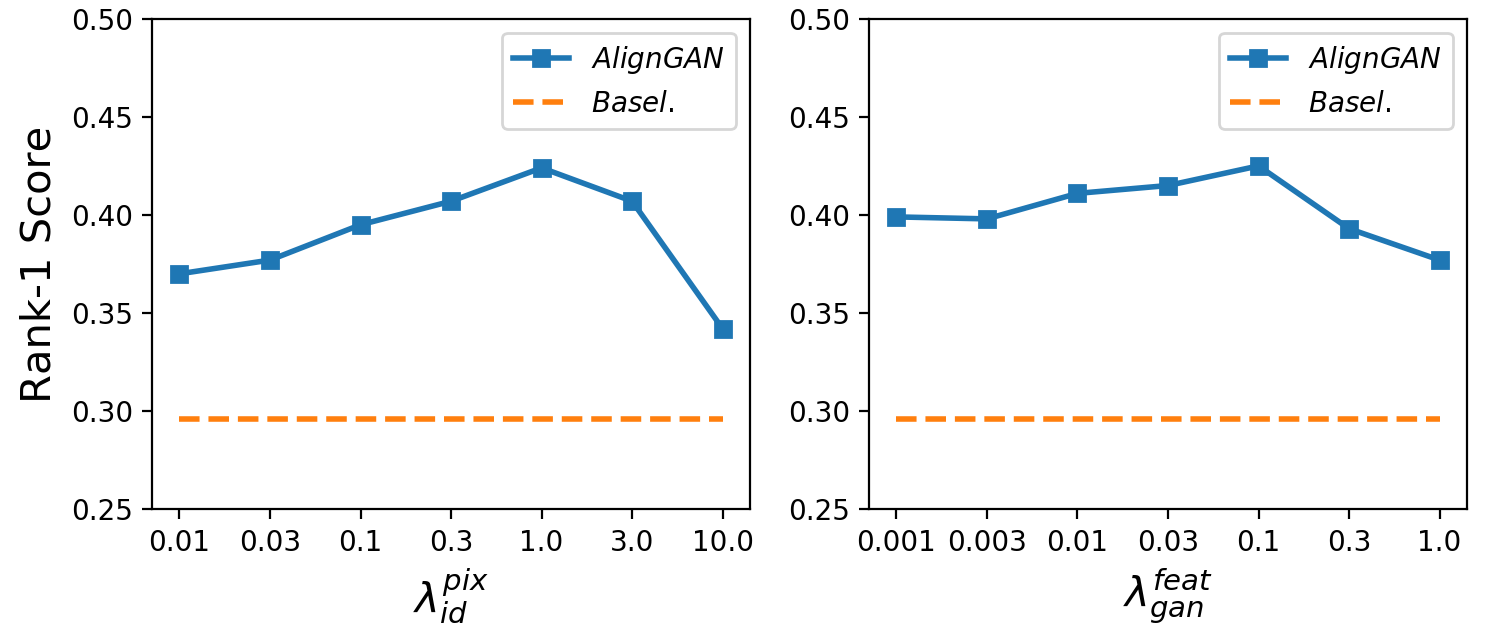}
\caption{
The results of parameter analysis for $\lambda^{pix}_{id}$ and $\lambda_{gan}^{feat}$.
}\label{fig:paramemter_analysis}
\end{figure}

%In this section, we give more results about ablation study, parameter analysis, and effect of different CNN Backbones.

\textbf{Ablation Study}. To evaluate each component of our $AlignGAN$ model, we conduct four variants with different settings.
Firstly, we straightly use real RGB and IR images to train the feature generator (\textit{i.e.} the CNN model) with the classification and triplet losses.
Secondly, to evaluate the pixel alignment module, we set $\lambda_{gan}^{feat}=0$ in Eq.(\ref{loss:feature_alignment_module}).
Thirdly, to evaluate the feature alignment module, we discard the pixel alignment module, thus the feature alignment module takes RGB and IR images as inputs.
Fourthly, to evaluate our joint discriminator module, we separately adopt the pixel and feature alignment modules with two discriminators as in Figure \ref{fig:compared_gans}(b).
We denote them as $Basel.$, $PixAlign$, $FeatAlign$ and $AlignGAN^-$  throughout the paper, respectively.

The experimental results are shown in Table \ref{table:ablation_study}.
Firstly, we can see that $PixAlign$ outperforms $Basel.$ by 11.0\% by Rank-1.
In addition, $FeatAlign$ outperforms $Basel.$ by 4.5\% Rank-1 score. These results demonstrate  the effectiveness of our pixel and feature alignment modules.
We also find that $PixAlign$ outperforms $FeatAlign$ by 6.5\% by Rank-1 score, which shows that the pixel alignment module plays a more important role than the feature alignment module.
Furthermore, $AlignGAN^{-}$ is comparable with $FeatAlign$ but worse than $PixAlign$, which implies that the simple stack of the two modules makes no contribution to better performance.
This may be because the cross-modality variation has been significantly reduced by the pixel alignment module. Also, the labels of training and test set are unshared. In the cases, enforcing align features by only fitting labels of training set cannot boost the performance too much, even import more noise and lead to identity-consistency.
Finally, the proposed model $AlignGAN$ performs much better than both $PixAlign$ and $FeatAlign$. This implies that our joint discriminator module can complement and enhance the pixel and feature alignment modules jointly.
More results will be discussed in Section \ref{section:visualization}.

\textbf{Parameters Analysis}. 
Here, we evaluate the effect of the  weights, \textit{i.e.} $\lambda^{pix}_{id}$ and $\lambda_{gan}^{feat}$.  As shown in Figure \ref{fig:paramemter_analysis}, we report the experimental results of our $AlignGAN$ with different $\lambda^{pix}_{id}$ and $\lambda_{gan}^{feat}$  on SYSU-MM01 dataset under the \textit{single-shot\&all-search} mode.
The $Basel.$ is defined in Section \ref{section:analysis}.
%which straightly uses RGB and IR images to train a CNN model with the classification and triplet losses.
It is clear that, when using different $\lambda^{pix}_{id}$ and $\lambda_{gan}^{feat}$, our $AlignGAN$ model stably outperforms $Basel.$.
The experimental results show that our $AlignGAN$ model is robust to different weights.

\subsection{Visualization of Learned Images and Features \label{section:visualization}}

\begin{figure}[h]
\center
\includegraphics[scale=0.41]{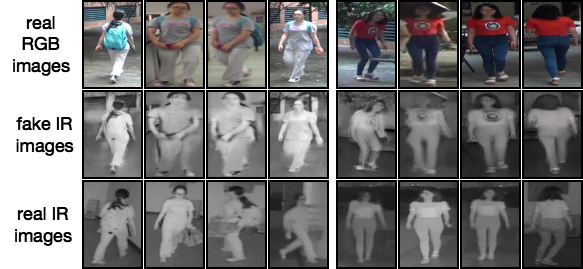}
\caption{
Fake IR images generated by our AlignGAN  (please view in color). 
The fake IR images can maintain identities and contents (such as views, poses) with original real RGB ones, and have the IR style. 
}
\label{fig:fake_ir_images}
\vspace{\VSpaceValue}
\end{figure}

\begin{figure}[h]
\center
\includegraphics[scale=0.20]{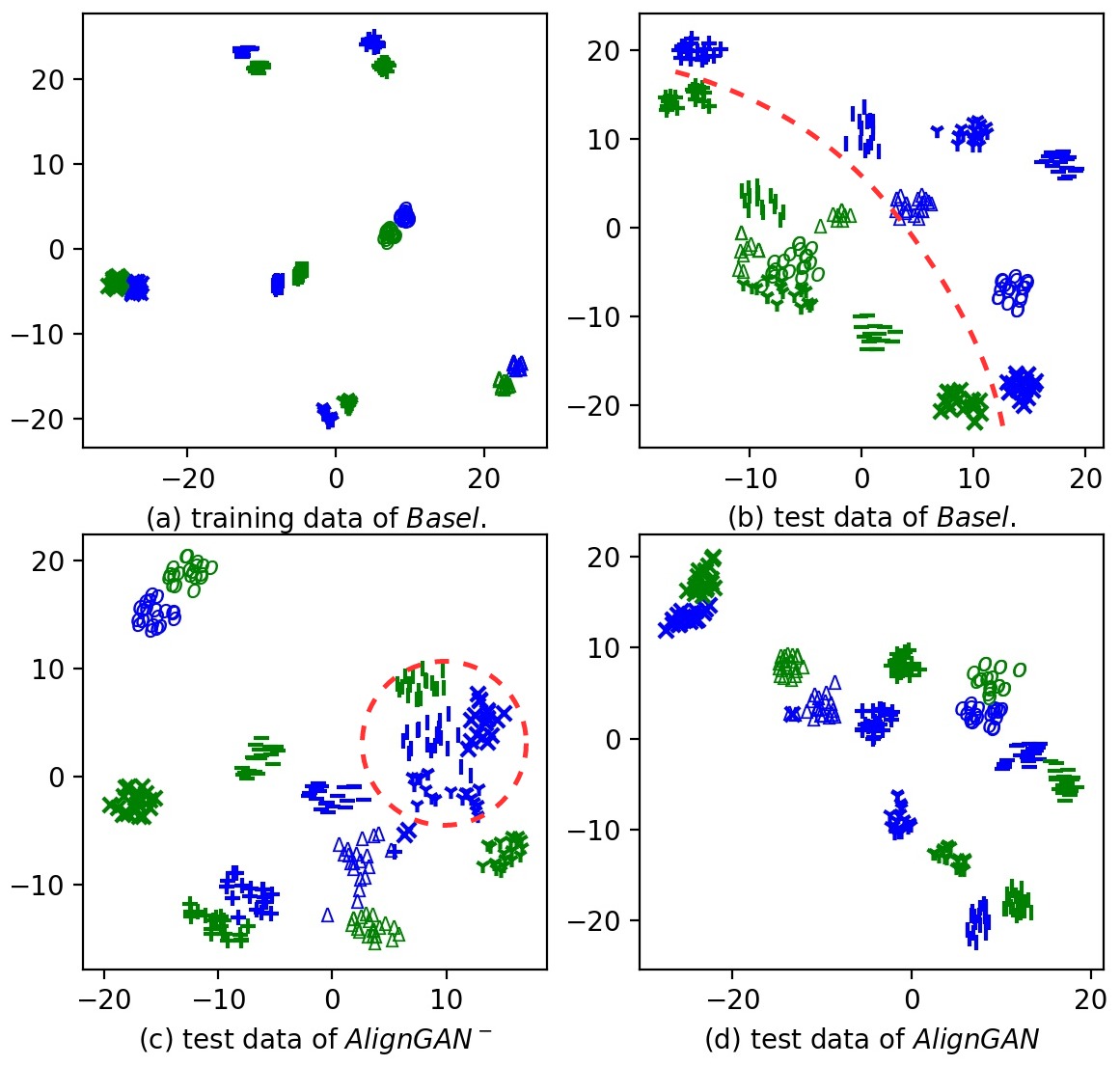}
\caption{
Visualization of learned features (please view in color).
Here, each color represents a modality, and each shape represents an identity. 
Features learned by AlignGAN can better maintain identity-consistency.
Please refer to the text for more details.
}
\label{fig:tsne}
\vspace{\VSpaceValue}
\end{figure}

To better understand the pixel and feature alignment modules, on SYSU-MM01 dataset, we display the fake IR images and the T-SNE \cite{maaten2008visualizing} distribution of the learned feature vectors in Figure {\ref{fig:fake_ir_images}} and Figure {\ref{fig:tsne}}, respectively.
As shown in Figure \ref{fig:fake_ir_images},  we can see that the fake IR images have similar contents (such as views, poses) and maintain identities of the corresponding real RGB images, meanwhile have the IR style. Thus, the generated fake IR images can bridge the gap between RGB and IR images, and the cross-modality variation in the pixel space can be reduced.

In Figure \ref{fig:tsne}, each color represents a modality, each shape represents an identity. The training data of $Basel.$ means feature vectors of training data learned with $Basel.$, and so on.
Note that $Basel.$ and $AlignGAN^-$ is defined in Section \ref{section:analysis}.
We have the following observations.
Firstly, when comparing Figure \ref{fig:tsne}(a) with Figure \ref{fig:tsne}(b), we can find $Basel.$ perfectly aligns two modalities of training data, while fails to do that for test data. As we can see in Figure \ref{fig:tsne}(b), the two modalities of test data can be easily separated by the red dotted line. This shows that it is difficult to reduce the cross-modality variation with a single feature alignment module.
Secondly, in Figure \ref{fig:tsne}(c), we can find that $AlignGAN^-$ performs much better modality-alignment than $Basel.$.
Even so, we can find the learned features fail  to maintain the identity-consistency, \textit{i.e.} some points are  aligned to wrong identities. For example, the red circle in Figure \ref{fig:tsne}(c) marks this case.
Finally, we can find our proposed model $AlignGAN$ as shown in Figure \ref{fig:tsne}(d) is able to not only reduce  the cross-modality variation, but also maintain the identity-consistency of features.
In summary, experimental results and analysis above demonstrate the effectiveness of $AlignGAN$.

\section{Conclusion}
In this paper, we propose a novel  Alignment Generative Adversarial Network  by
exploiting  pixel alignment and feature alignment jointly for the RGB-IR RE-ID task.
The proposed model is able to not only alleviate the cross-modality and
intra-modality variations, but also learn identity-consistent
features.
Extensive experimental results on two standard
benchmarks demonstrate that our  model
performs favorably against state-of-the-art methods.

\section*{Acknowledgments}

This work was supported in part by the National Key Research and Development Program 2018YFB0804204, National Natural Science Foundation of China under Grants 61720106012, 61533016, 61806203, 61572498, 61751211 and 61728210, the Strategic Priority Research Program of Chinese Academy of Science under Grant XDBS01000000 and XDB32050200, the Beijing Natural Science Foundation under Grant L172050 and 4172062, and Youth Innovation Promotion Association CAS2018166.

{\small
\bibliographystyle{ieee_fullname}
\bibliography{egbib}

\begin{thebibliography}{10}\itemsep=-1pt

\bibitem{choi2018stargan}
Y.~{Choi}, M.~{Choi}, M.~{Kim}, J.-W. {Ha}, S.~{Kim}, and J.~{Choo}.
\newblock Stargan: Unified generative adversarial networks for multi-domain
  image-to-image translation.
\newblock In {\em 2018 IEEE/CVF Conference on Computer Vision and Pattern
  Recognition}, pages 8789--8797, 2018.

\bibitem{dai2018cross}
P.~{Dai}, R.~{Ji}, H.~{Wang}, Q.~{Wu}, and Y.~{Huang}.
\newblock Cross-modality person re-identification with generative adversarial
  training.
\newblock In {\em IJCAI 2018: 27th International Joint Conference on Artificial
  Intelligence}, pages 677--683, 2018.

\bibitem{dalal2005histograms}
N.~Dalal and B.~Triggs.
\newblock Histograms of oriented gradients for human detection.
\newblock In {\em international Conference on computer vision \& Pattern
  Recognition (CVPR'05)}, volume~1, pages 886--893. IEEE Computer Society,
  2005.

\bibitem{dou2018unsupervised}
Q.~{Dou}, C.~{Ouyang}, C.~{Chen}, H.~{Chen}, and P.-A. {Heng}.
\newblock Unsupervised cross-modality domain adaptation of convnets for
  biomedical image segmentations with adversarial loss.
\newblock In {\em IJCAI 2018: 27th International Joint Conference on Artificial
  Intelligence}, pages 691--697, 2018.

\bibitem{ganin2016domain}
Y.~{Ganin}, E.~{Ustinova}, H.~{Ajakan}, P.~{Germain}, H.~{Larochelle},
  F.~{Laviolette}, M.~{Marchand}, and V.~S. {Lempitsky}.
\newblock Domain-adversarial training of neural networks.
\newblock {\em Journal of Machine Learning Research}, 17(1):1--35, 2016.

\bibitem{gong2014person}
S.~{Gong}, M.~{Cristani}, S.~{Yan}, and C.~C. {Loy}.
\newblock {\em Person Re-Identification}.
\newblock 2014.

\bibitem{goodfellow2014generative}
I.~J. {Goodfellow}, J.~{Pouget-Abadie}, M.~{Mirza}, B.~{Xu}, D.~{Warde-Farley},
  S.~{Ozair}, A.~C. {Courville}, and Y.~{Bengio}.
\newblock Generative adversarial nets.
\newblock In {\em Advances in Neural Information Processing Systems 27}, pages
  2672--2680, 2014.

\bibitem{hao2019hsme}
Y.~{Hao}, N.~{Wang}, J.~{Li}, and X.~{Gao}.
\newblock Hsme hypersphere manifold embedding for visible thermal person
  re-identification.
\newblock In {\em AAAI-19 AAAI Conference on Artificial Intelligence}, 2019.

\bibitem{he2016deep}
K.~{He}, X.~{Zhang}, S.~{Ren}, and J.~{Sun}.
\newblock Deep residual learning for image recognition.
\newblock In {\em 2016 IEEE Conference on Computer Vision and Pattern
  Recognition (CVPR)}, pages 770--778, 2016.

\bibitem{hermans2017defense}
A.~{Hermans}, L.~{Beyer}, and B.~{Leibe}.
\newblock In defense of the triplet loss for person re-identification.
\newblock {\em arXiv preprint arXiv:1703.07737}, 2017.

\bibitem{hoffman2018cycada}
J.~{Hoffman}, E.~{Tzeng}, T.~{Park}, J.-Y. {Zhu}, P.~{Isola}, K.~{Saenko},
  A.~A. {Efros}, and T.~{Darrell}.
\newblock Cycada: Cycle-consistent adversarial domain adaptation.
\newblock {\em international conference on machine learning}, pages 1989--1998,
  2018.

\bibitem{ioffe2015batch}
S.~{Ioffe} and C.~{Szegedy}.
\newblock Batch normalization: Accelerating deep network training by reducing
  internal covariate shift.
\newblock {\em international conference on machine learning}, pages 448--456,
  2015.

\bibitem{isola2017image}
P.~{Isola}, J.-Y. {Zhu}, T.~{Zhou}, and A.~A. {Efros}.
\newblock Image-to-image translation with conditional adversarial networks.
\newblock In {\em 2017 IEEE Conference on Computer Vision and Pattern
  Recognition (CVPR)}, pages 5967--5976, 2017.

\bibitem{wu2019clustering}
Z.~L. X. W. Y.~Y. Jinlin~Wu, Shengcai~Liao and S.~Z. Li.
\newblock Clustering and dynamic sampling for unsupervised domain adaptation in
  person re-identification.
\newblock In {\em ICME 2019}. IEEE, 2019.

\bibitem{koestinger2012large}
M.~Koestinger, M.~Hirzer, P.~Wohlhart, P.~M. Roth, and H.~Bischof.
\newblock Large scale metric learning from equivalence constraints.
\newblock In {\em 2012 IEEE conference on computer vision and pattern
  recognition}, pages 2288--2295. IEEE, 2012.

\bibitem{liao2015person}
S.~Liao, Y.~Hu, X.~Zhu, and S.~Z. Li.
\newblock Person re-identification by local maximal occurrence representation
  and metric learning.
\newblock In {\em Proceedings of the IEEE conference on computer vision and
  pattern recognition}, pages 2197--2206, 2015.

\bibitem{liao2015efficient}
S.~Liao and S.~Z. Li.
\newblock Efficient psd constrained asymmetric metric learning for person
  re-identification.
\newblock In {\em Proceedings of the IEEE International Conference on Computer
  Vision}, pages 3685--3693, 2015.

\bibitem{luo2019bag}
H.~Luo, Y.~Gu, X.~Liao, S.~Lai, and W.~Jiang.
\newblock Bag of tricks and a strong baseline for deep person
  re-identification.
\newblock In {\em Proceedings of the IEEE Conference on Computer Vision and
  Pattern Recognition Workshops}, 2019.

\bibitem{ma2014covariance}
B.~Ma, Y.~Su, and F.~Jurie.
\newblock Covariance descriptor based on bio-inspired features for person
  re-identification and face verification.
\newblock {\em Image and Vision Computing}, 32(6-7):379--390, 2014.

\bibitem{mirza2014conditional}
M.~{Mirza} and S.~{Osindero}.
\newblock Conditional generative adversarial nets.
\newblock {\em arXiv preprint arXiv:1411.1784}, 2014.

\bibitem{nguyen2017person}
D.~T. {Nguyen}, H.~G. {Hong}, K.-W. {Kim}, and K.~R. {Park}.
\newblock Person recognition system based on a combination of body images from
  visible light and thermal cameras.
\newblock {\em Sensors}, 17(3):605, 2017.

\bibitem{radford2016unsupervised}
A.~{Radford}, L.~{Metz}, and S.~{Chintala}.
\newblock Unsupervised representation learning with deep convolutional
  generative adversarial networks.
\newblock {\em international conference on learning representations}, 2016.

\bibitem{russakovsky2015imagenet}
O.~{Russakovsky}, J.~{Deng}, H.~{Su}, J.~{Krause}, S.~{Satheesh}, S.~{Ma},
  Z.~{Huang}, A.~{Karpathy}, A.~{Khosla}, M.~S. {Bernstein}, A.~C. {Berg}, and
  L.~{Fei-Fei}.
\newblock Imagenet large scale visual recognition challenge.
\newblock {\em International Journal of Computer Vision}, 115(3):211--252,
  2015.

\bibitem{shao2018feature}
R.~{Shao}, X.~{Lan}, and P.~C. {Yuen}.
\newblock Feature constrained by pixel: Hierarchical adversarial deep domain
  adaptation.
\newblock In {\em Proceedings of the 26th ACM international conference on
  Multimedia}, pages 220--228, 2018.

\bibitem{sun2018beyond}
Y.~Sun, L.~Zheng, Y.~Yang, Q.~Tian, and S.~Wang.
\newblock Beyond part models: Person retrieval with refined part pooling (and a
  strong convolutional baseline).
\newblock In {\em Proceedings of the European Conference on Computer Vision
  (ECCV)}, pages 480--496, 2018.

\bibitem{maaten2008visualizing}
L.~van~der {Maaten} and G.~E. {Hinton}.
\newblock Visualizing data using t-sne.
\newblock {\em Journal of Machine Learning Research}, 9:2579--2605, 2008.

\bibitem{guan2019color}
G.~{Wang}, Y.~{Yang}, J.~{Cheng}, J.~{Wang}, and Z.~{Hou}.
\newblock Color-sensitive person re-identification.
\newblock In {\em IJCAI 2019: 28th International Joint Conference on Artificial
  Intelligence}, 2019.

\bibitem{wu2017rgb}
A.~{Wu}, W.-S. {Zheng}, H.-X. {Yu}, S.~{Gong}, and J.~{Lai}.
\newblock Rgb-infrared cross-modality person re-identification.
\newblock In {\em 2017 IEEE International Conference on Computer Vision
  (ICCV)}, pages 5390--5399, 2017.

\bibitem{yang2016large}
Y.~{Yang}, S.~{Liao}, Z.~{Lei}, and S.~Z. {Li}.
\newblock Large scale similarity learning using similar pairs for person
  verification.
\newblock In {\em AAAI'16 Proceedings of the Thirtieth AAAI Conference on
  Artificial Intelligence}, pages 3655--3661, 2016.

\bibitem{yang2014salient}
Y.~Yang, J.~Yang, J.~Yan, S.~Liao, D.~Yi, and S.~Z. Li.
\newblock Salient color names for person re-identification.
\newblock In {\em European conference on computer vision}, pages 536--551.
  Springer, 2014.

\bibitem{ye2018hierarchical}
M.~{Ye}, X.~{Lan}, J.~{Li}, and P.~c~{Yuen}.
\newblock Hierarchical discriminative learning for visible thermal person
  re-identification.
\newblock In {\em AAAI-18 AAAI Conference on Artificial Intelligence}, pages
  7501--7508, 2018.

\bibitem{ye2018visible}
M.~{Ye}, Z.~{Wang}, X.~{Lan}, and P.~C. {Yuen}.
\newblock Visible thermal person re-identification via dual-constrained
  top-ranking.
\newblock In {\em IJCAI 2018: 27th International Joint Conference on Artificial
  Intelligence}, pages 1092--1099, 2018.

\bibitem{zhang2018correlation}
T.~{Zhang}, S.~{Liu}, C.~{Xu}, B.~{Liu}, and M.-H. {Yang}.
\newblock Correlation particle filter for visual tracking.
\newblock {\em IEEE Transactions on Image Processing}, 27(6):2676--2687, 2018.

\bibitem{zhang2019learning}
T.~{Zhang}, C.~{Xu}, and M.-H. {Yang}.
\newblock Learning multi-task correlation particle filters for visual tracking.
\newblock {\em IEEE Transactions on Pattern Analysis and Machine Intelligence},
  41(2):365--378, 2019.

\bibitem{zhang2019robust}
T.~{Zhang}, C.~{Xu}, and M.-H. {Yang}.
\newblock Robust structural sparse tracking.
\newblock {\em IEEE Transactions on Pattern Analysis and Machine Intelligence},
  41(2):473--486, 2019.

\bibitem{zheng2016person}
L.~Zheng, Y.~Yang, and A.~G. Hauptmann.
\newblock Person re-identification: Past, present and future.
\newblock {\em arXiv preprint arXiv:1610.02984}, 2016.

\bibitem{zheng2013reidentification}
W.-S. Zheng, S.~Gong, and T.~Xiang.
\newblock Reidentification by relative distance comparison.
\newblock {\em IEEE transactions on pattern analysis and machine intelligence},
  35(3):653--668, 2013.

\bibitem{zheng2019joint}
Z.~Zheng, X.~Yang, Z.~Yu, L.~Zheng, Y.~Yang, and J.~Kautz.
\newblock Joint discriminative and generative learning for person
  re-identification.
\newblock {\em IEEE Conference on Computer Vision and Pattern Recognition
  (CVPR)}, 2019.

\bibitem{zhu2017unpaired}
J.-Y. {Zhu}, T.~{Park}, P.~{Isola}, and A.~A. {Efros}.
\newblock Unpaired image-to-image translation using cycle-consistent
  adversarial networks.
\newblock In {\em 2017 IEEE International Conference on Computer Vision
  (ICCV)}, pages 2242--2251, 2017.

\end{thebibliography}
}

\end{document}